\documentclass[runningheads]{llncs}
\usepackage[T1]{fontenc}
\usepackage[utf8]{inputenc}
\usepackage{subcaption}
\usepackage{float}
\usepackage{tabularx}
\usepackage{mdframed}
\usepackage{array}
\usepackage[table,xcdraw]{xcolor} % Loaded once with options
\usepackage{multirow}
\usepackage{adjustbox}
\usepackage{amssymb}
\usepackage{caption}
\usepackage{booktabs}
\usepackage{soul}
\usepackage{pdfpages}
\usepackage{amsmath}
\usepackage[numbers,sort&compress]{natbib}
\usepackage{fancyhdr}
\usepackage{hyperref} % hyperref should usually be loaded last

\usepackage{acronym}
\newacro{gan}[GAN]{Generative Adversarial Network}
\newacro{vlm}[VLM]{Vision-Language Model}
\newacro{dm}[DM]{Diffusion Model}
\newacro{ai}[AI]{Artificial Intelligence}
\newacro{cnn}[CNN]{Convolutional Neural Network}
\newacro{vit}[ViT]{Vision Transformer}
\newacro{vae}[VAE]{Variational Autoencoder}
\newacro{ic}[IC]{Image Captioning}
\newacro{vqa}[VQA]{Visual Question Answering}
\newacro{nlp}[NLP]{Natural Language Processing}
\newacro{cv}[CV]{Computer Vision}
\newacro{t2i}[T2I]{text-to-image}
\newacro{pg}[PG]{Photographs}
\newacro{srm}[SRM]{spatial rich model}
\newacro{svm}[SVM]{support vector machine}
\newacro{dct}[DCT]{discrete cosine transform}
\newacro{aide}[AIDE]{(AI-generated Image Detector with Hybrid Features)}
\newacro{mlp}[MLP]{multi layer preservon}
\newacro{dnn}[DNN]{Deep Neural Network}

\usepackage{graphicx} % For scaling the table
\usepackage[table]{xcolor} % For cell coloring
\usepackage{booktabs} % For professional looking tables
\usepackage{xcolor}
\usepackage{soul}
\usepackage{pdfpages}

\usepackage{amsmath}
\usepackage[numbers,sort&compress]{natbib}
\usepackage{fancyhdr}
\newmdenv[
    backgroundcolor=yellow,
    linecolor=yellow,
    innertopmargin=10pt,
    innerbottommargin=10pt,
    innerleftmargin=10pt,
    innerrightmargin=10pt
]{highlighted}

\begin{document}

\title{SPARK-IL: Spectral Retrieval-Augmented RAG for Knowledge-driven Deepfake Detection via Incremental Learning}

\titlerunning{Synthetic Image Detection using  Distillation and VLMs}
\author{Hessen Bougueffa Eutamene \inst{1} \and
Abdellah  Zakaria Sellam\inst{2,3} \and
Abdelmalik Taleb-Ahmed\inst{1}\and
Abdenour Hadid\inst{4}}
\authorrunning{Eutamene {\it et al.}}
%\institute{Laboratory of IEMN, CNRS, Centrale Lille, UMR 8520, Univ. Polytechnique Hauts-de-France, Valenciennes, F-59313, France  
\institute{Univ. Polytechnique Hauts-de-France, Valenciennes, France
\\
\and  
%Institute of Applied Sciences and Intelligent Systems – CNR, Via per Monteroni, 73100 Lecce, Lecce, Italy
Institute of Applied Sciences and Intelligent Systems – Lecce, Italy
\\ 
\and
%Department of Innovation Engineering, University of Salento \& Institute of Applied Sciences and Intelligent Systems – CNR, Via per Monteroni, 73100 Lecce, Lecce, Italy\\
Department of Innovation Engineering, University of Salento, Italy\\
 \and
Sorbonne Center for Artificial Intelligence, Sorbonne University Abu Dhabi\\
}
%\includepdf[pages=-]{Reb3.pdf}
\maketitle              
\begin{abstract}

Detecting AI-generated images remains challenging, as detectors trained on specific generators often fail to generalize to unseen models. While pixel-level artifacts vary across models, frequency-domain signatures exhibit greater consistency, providing a promising foundation for cross-generator detection.
We propose SPARK-IL, a retrieval-augmented framework combining dual-path spectral analysis with incremental learning. A partially frozen ViT-L/14 encoder extracts semantic representations, while a parallel path embeds raw RGB pixels. Both undergo multi-band Fourier decomposition into four frequency bands, each processed by Kolmogorov –Arnold Networks (KAN) with mixture-of-experts for band-specific transformations. Spectral embeddings are fused via cross-attention with residual connections.
During inference, the fused embedding retrieves $k$ nearest labeled signatures from a Milvus database using cosine similarity, with predictions via majority voting. Incremental learning expands the database while elastic weight consolidation preserves previously learned transformations.
On the UniversalFakeDetect benchmark (19 generative models including GANs, face-swapping, and diffusion methods), SPARK-IL achieves 94.6\% mean accuracy. Code will be publicly released at:
https://github.com/HessenUPHF/SPARK-IL.

\keywords{Deepfake detection\and
Retrieval-augmented generation\and
Multi-band spectral analysis\and
Kolmogorov-Arnold Networks\and
Dual-path architecture\and
Incremental learning}

\end{abstract}

\section{Introduction}

\begin{figure}[h]
    \centering
    \includegraphics[width=1.0\linewidth]{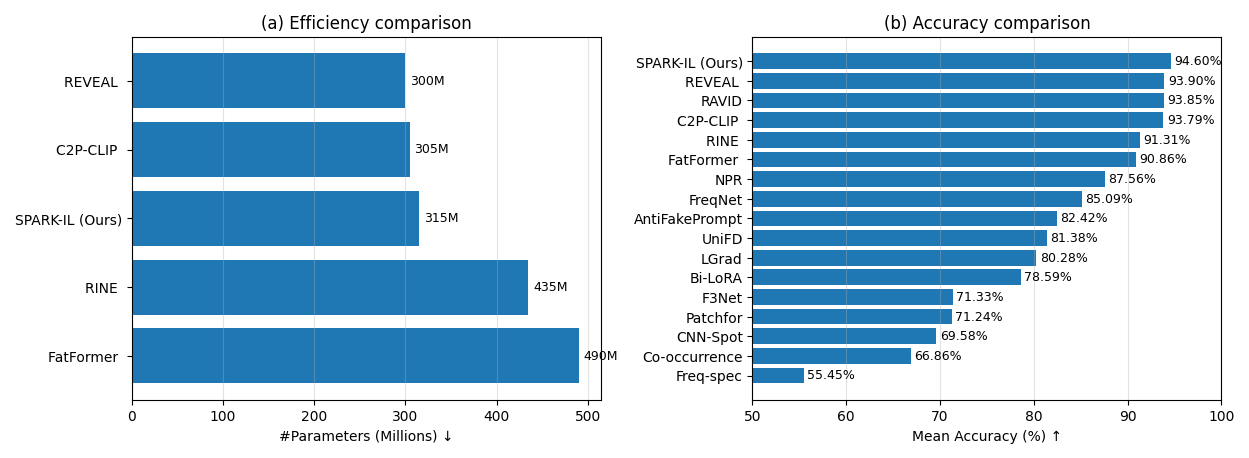}
    \caption{Comparison of detection accuracy and model efficiency on the UniversalFakeDetect benchmark. 
    (a) Efficiency comparison in terms of number of parameters. 
    (b) Mean accuracy across 19 generative models.}
    \label{fig:main_results}
\end{figure}

Recent advances in generative modeling have enabled the synthesis of highly realistic images that are increasingly difficult to distinguish from authentic content. Modern generative adversarial networks (GANs) and diffusion-based models now closely approximate the statistical properties of natural images, rendering visual inspection unreliable and posing significant challenges for automated detection systems \cite{NIPS2014_f033ed80,dhariwal2021diffusion}. While these models enable beneficial applications, they also facilitate malicious uses such as misinformation, identity impersonation, and content manipulation.
Despite extensive research efforts, current deepfake detection methods remain fragile under real-world conditions. Most approaches rely on generator-specific spatial artifacts or architectural fingerprints that are effective in closed-world settings but fail to generalize to unseen synthesis pipelines \cite{Wang_2020_CVPR,ojha2023fakedetect}. As generative models continue to suppress visible artifacts and improve perceptual realism, the assumption that such cues transfer across generators becomes increasingly unreliable.
A key observation motivating this work is that, although pixel-level artifacts vary substantially across generative models, frequency-domain statistics exhibit greater consistency. Synthetic images tend to deviate from natural spectral distributions in characteristic ways that persist across generator families. These deviations are often weak or invisible in the spatial domain but become more pronounced when analyzed in the frequency domain \cite{10.1007/978-3-030-58610-2_6,10.1609/aaai.v38i5.28310}.
Another fundamental limitation of existing detectors is their static nature. Once trained, parametric classifiers lack mechanisms to incorporate knowledge about newly emerging generators without retraining from scratch, often leading to catastrophic forgetting \cite{doi:10.1073/pnas.1611835114}. In contrast, retrieval-based inference provides a non-parametric alternative in which predictions are grounded in similarity to previously observed examples rather than fixed decision boundaries \cite{lewis2020retrieval}. This paradigm naturally supports open-world settings and enables adaptive behavior under distribution shift.
We propose SPARK-IL, a retrieval-augmented framework for incremental deepfake detection combining dual-path spectral analysis with frequency-aware embedding retrieval. The method extracts complementary spectral representations from raw RGB pixels and high-level semantic features using multi-band Fourier analysis. Band-specific Kolmogorov–Arnold Networks (KAN) model nonlinear frequency transformations, fused via cross-attention to capture pixel-level and feature-level spectral interactions. At inference, predictions are obtained through nearest-neighbor retrieval and majority voting over a continuously expanding database. Incremental adaptation to new generators is achieved via elastic weight consolidation, enabling knowledge accumulation without degrading performance on previously seen techniques. As shown in Figure~\ref{fig:main_results}, SPARK-IL achieves superior detection accuracy while maintaining competitive model efficiency compared to recent state-of-the-art methods.\\

\noindent\textbf{Main Contributions:} The main contributions of this paper include:
\vspace{-1mm}
\begin{itemize}
\item A dual-path multi-band spectral architecture processing pixel-level and feature-level representations via parallel Fast Fourier Transform (FFT) blocks, with KAN modules learning band-specific transformations fused through cross-attention.
    
\item Retrieval-augmented classification via cosine similarity search and majority voting over spectral signatures in a Milvus database.
    
\item Incremental learning via elastic weight consolidation and continuous embedding indexing, enabling adaptation to new generators without forgetting.
    
\item Evaluation on 19 generative models achieving 94.60\% mean accuracy with consistent cross-category generalization.
\end{itemize}
\vspace{0.75em}

The remainder of this paper is organized as follows. Section~\ref{RW} reviews related work on synthetic image generation and detection, with an emphasis on spectral-based and retrieval-augmented approaches. Section~\ref{sec:methodology} presents the proposed SPARK-IL framework, including the dual spectral architecture, multi-band KAN-FFT processing, and retrieval-augmented inference. Section~\ref{EXS} describes the experimental setup. Section~\ref{RES} reports quantitative and qualitative results, including the ablation study. Section~\ref{sec:discussion} discusses the implications and limitations of the proposed approach, and Section~\ref{sec:conc} concludes the paper and outlines future research directions.

%\end{itemize}
\vspace{-10mm}
\section{Related Work}
\label{RW}

Deep generative models have evolved rapidly. GANs \cite{NIPS2014_f033ed80} and their extensions \cite{karras2018progressive} enabled realistic synthesis but suffer from training instability. Diffusion models now dominate: latent diffusion \cite{Rombach_2022_CVPR} underpins GLIDE \cite{nichol2022glide}, Imagen \cite{saharia2022photorealistic}, and DALL·E~2/3 \cite{ramesh2021zero,openai2023dalle3}, while transformer-based and hybrid generators \cite{tao2023galip,Xu2024UFOGen} further expand the landscape.

This evolution complicates detection as generator-specific artifacts are increasingly suppressed \cite{10.1145/3746265.3759667}. While spatial artifacts vary across models, frequency-domain characteristics exhibit greater consistency, motivating spectral analysis for cross-generator detection.

Retrieval-Augmented Generation (RAG) enhances predictions by incorporating external knowledge at inference \cite{lewis2020retrieval}. Extended from NLP to vision, early approaches relied on nearest-neighbor matching but exhibited limited robustness under distribution shift \cite{koutlis2024leveraging}. Recent work integrates retrieval into vision-language models: multimodal context encoding \cite{long2025retrievalaugmentedvisualquestionanswering}, structured knowledge retrieval \cite{Caffagni2024WikiLLaVA}, inference-time generalization \cite{bonomo2025visualragexpandingmllm}, and graph-based approaches \cite{NEURIPS2024_efaf1c97}.

For synthetic detection, retrieval enables decisions based on similarity to labeled exemplars rather than fixed boundaries, improving robustness to unseen generators \cite{RAVID}. However, existing approaches rely on semantic or pixel-level features that vary across generators. Our method retrieves spectral embeddings capturing frequency-domain characteristics and employs majority voting over $k$ nearest neighbors, reducing outlier sensitivity.

Early CNN methods identify spatial artifacts (texture inconsistencies, noise patterns) \cite{Wang_2020_CVPR} but exhibit limited robustness to unseen generators and post-processing, as spatial artifacts vary across synthesis pipelines. Frequency-domain analysis captures generator fingerprints invisible in pixel space \cite{10860248}, yet most methods operate only on raw pixels and remain sensitive to compression and resizing, missing artifacts that manifest differently across representation levels.
Transformer-based detectors improve global representation but suffer from domain-shift sensitivity \cite{liu2024fat}. Reconstruction-based approaches detect synthetic content through reconstruction error discrepancies \cite{wang2023dire} but introduce inference overhead and rely on fixed parametric models that cannot readily adapt to new generators.
Vision-language approaches based on CLIP embeddings capture semantic inconsistencies and can be enhanced through prompt-based strategies \cite{cozzolino2023raising,tan2024c2pclipinjectingcategorycommon}. More recent methods formulate detection as visual question answering to provide textual explanations \cite{10.1111/exsy.13829}. Nevertheless, these approaches depend on pre-trained representations not explicitly designed for spectral artifact modeling.
Overall, existing methods are limited by their reliance on single-level representations, static decision boundaries, and poor adaptability to emerging generators. SPARK-IL addresses these challenges through dual-path multi-band spectral analysis, retrieval-augmented inference via majority voting, and incremental database expansion with elastic weight consolidation.

\section{Proposed Method}
\label{sec:methodology}

We present SPARK-IL, a spectral analysis framework for deepfake detection combining dual-path frequency processing with retrieval-augmented inference. The method exploits distinctive spectral signatures at both pixel and semantic levels that persist across generation techniques.
\subsection{Overview}
\begin{figure}[h]
    \centering
    \includegraphics[width=0.85\linewidth]{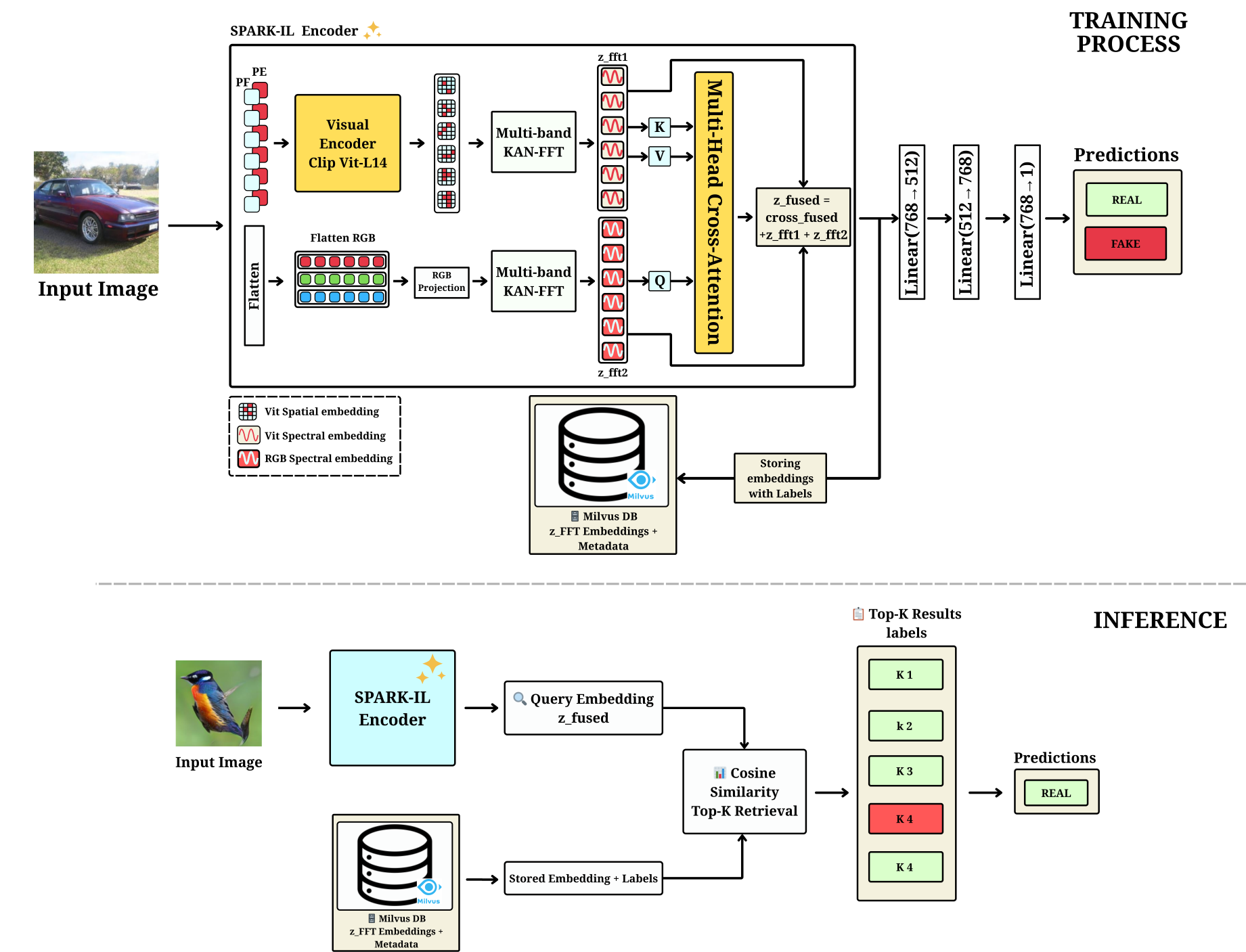}
    \caption{SPARK-IL architecture. Dual spectral paths process ViT features and RGB pixels through multi-band KAN-FFT blocks. Cross-attention fuses the spectral embeddings, which are used for classification during training and stored in Milvus for retrieval-based inference.}
    \label{fig:Archi}
\end{figure}

Figure~\ref{fig:Archi} illustrates the architecture. Two parallel paths extract features: a ViT encoder captures semantic representations, while a projection layer maps raw RGB pixels to the same embedding dimension. Both undergo multi-band FFT decomposition and KAN-based processing. The resulting spectral embeddings are fused via cross-attention with residual connections, serving parametric classification during training and retrieval-augmented prediction at inference.

\setlength{\textfloatsep}{6pt}   
\setlength{\intextsep}{6pt}      
\setlength{\floatsep}{6pt}       
\setlength{\abovecaptionskip}{3pt}
\setlength{\belowcaptionskip}{3pt}

The semantic path employs a partially frozen ViT-L/14 encoder, where blocks 0--21 remain fixed to preserve pre-trained knowledge while blocks 22--23 are fine-tuned for task adaptation. The encoder produces an embedding $\mathbf{h}_{\text{vit}} \in \mathbb{R}^{d}$ with $d=768$. In parallel, the pixel path flattens the input image and projects it through a linear layer to obtain $\mathbf{h}_{\text{rgb}} \in \mathbb{R}^{d}$. This dual-path design enables the framework to capture both high-level semantic inconsistencies and low-level pixel artifacts that characterize synthetic images.

\subsection{Multi-Band Spectral Processing}

\setlength{\textfloatsep}{6pt}  
\setlength{\intextsep}{6pt}      
\setlength{\floatsep}{6pt}       
\setlength{\abovecaptionskip}{3pt}
\setlength{\belowcaptionskip}{3pt}

\begin{figure}[h]
    \centering
    \includegraphics[width=0.9\linewidth]{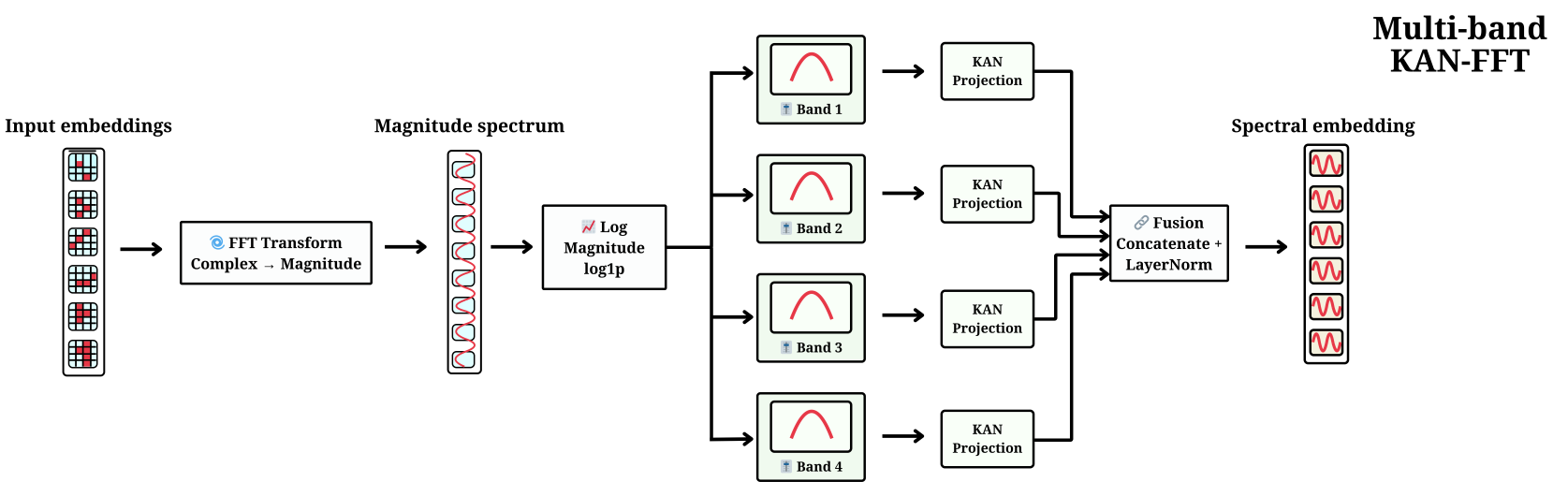}
    \caption{Multi-band KAN-FFT block. Input features are transformed via FFT, converted to log-magnitude spectrum, partitioned into four frequency bands, processed through dedicated KAN layers, and fused into a spectral embedding.}
    \label{fig:Multi}
\end{figure}

The core component of SPARK-IL is the Multi-Band KAN-FFT block (Figure~\ref{fig:Multi}), which transforms features into the frequency domain and processes different spectral ranges independently. This design exploits the observation that deepfake artifacts manifest at distinct frequency bands: low-frequency components capture structural and lighting inconsistencies, while high-frequency components reveal generation-specific noise patterns.

The block first applies 1D FFT to the input feature vector and computes the log-magnitude spectrum:
\begin{equation}
\mathbf{h}_{\text{freq}} = \log(1 + |\mathcal{F}(\mathbf{h})|) \in \mathbb{R}^{d}
\end{equation}

The spectrum is then partitioned into four disjoint bands of dimension $d_b = d/4 = 192$, covering frequency ranges $[0, 0.25\pi]$, $[0.25\pi, 0.5\pi]$, $[0.5\pi, 0.75\pi]$, and $[0.75\pi, \pi]$. Each band is processed through a Kolmogorov-Arnold Network (KAN) with mixture-of-experts:
\begin{equation}
\mathbf{h}_{\text{kan}}^{(b)} = \mathbf{W}^{(b)} \mathbf{h}_{\text{freq}}^{(b)} + \sum_{e=1}^{E} \alpha_e^{(b)} \cdot \text{Expert}_e^{(b)}(\mathbf{h}_{\text{freq}}^{(b)})
\end{equation}
where $E=4$ experts provide band-specific nonlinear transformations and the gating weights $\alpha_e^{(b)}$ are computed via input-dependent softmax routing. The band outputs are concatenated and projected to form the final spectral embedding. Each path employs a stack of two Multi-Band KAN-FFT blocks with intermediate normalization layers.

\subsection{Cross-Attention Fusion}

The dual spectral embeddings $\mathbf{z}_{\text{fft1}}$ (from RGB path) and $\mathbf{z}_{\text{fft2}}$ (from ViT path) are fused via multi-head cross-attention with 12 heads. The RGB spectral embedding serves as the query while the ViT spectral embedding provides keys and values, enabling the model to identify correlations between pixel-level spectral anomalies and semantic-level inconsistencies:
\begin{equation}
\mathbf{h}_{\text{cross}} = \text{MultiHeadCrossAttn}(\mathbf{z}_{\text{fft1}}, \mathbf{z}_{\text{fft2}}) + \mathbf{z}_{\text{fft1}}
\end{equation}

The final fused representation incorporates weighted residual connections from both spectral paths:
\begin{equation}
\mathbf{h}_{\text{fused}} = \mathbf{h}_{\text{cross}} + 0.2 \cdot \mathbf{z}_{\text{fft1}} + 0.2 \cdot \mathbf{z}_{\text{fft2}}
\end{equation}

\subsection{Retrieval-Augmented Inference}

During training, $\mathbf{h}_{\text{fused}}$ passes through a projection head and linear classifier optimized with binary cross-entropy loss. Fused embeddings are stored in a Milvus database with ground-truth labels and generator identifiers.
At inference, the system extracts the test embedding, retrieves the $K{=}5$ nearest neighbors via cosine similarity, and classifies via majority voting over retrieved labels:
\begin{equation}
y_{\text{pred}} = \begin{cases}
0 \text{ (real)} & \text{if } \sum_{k=1}^{K} \mathbb{1}[y_{k} = 0] > K/2 \\
1 \text{ (fake)} & \text{otherwise}
\end{cases}
\end{equation}

This retrieval mechanism provides evidence-based predictions that adapt to the local query distribution without requiring parameter updates, enabling generalization to unseen generators.

\subsection{Incremental Learning}

To integrate new generators without catastrophic forgetting, SPARK-IL combines three mechanisms. Experience replay maintains a buffer of representative samples from previous techniques, mixed with current training data. Knowledge distillation regularizes the model by minimizing the distance between current and previous embeddings and logits:
\begin{equation}
\mathcal{L}_{\text{distill}} = \lambda_{\text{emb}} \|\mathbf{h}_{\text{fused}} - \mathbf{h}_{\text{fused}}^{\text{teacher}}\|_2^2 + \lambda_{\text{logit}} \|\hat{y} - \hat{y}^{\text{teacher}}\|_2^2
\end{equation}

Third, elastic weight consolidation penalizes parameter drift from the previous model state:
\begin{equation}
\mathcal{L}_{\text{EWC}} = \lambda_{\text{reg}} \|\boldsymbol{\theta} - \boldsymbol{\theta}^{\text{prev}}\|_2^2
\end{equation}

The total training objective combines classification loss with both regularization terms. After each incremental training phase, new embeddings are indexed in the Milvus database, expanding the retrieval knowledge base to cover the newly encountered generation technique.

\section{Experimental Setup}
\label{EXS}

\subsection{Dataset}

All experiments are conducted on the \textbf{UniversalFakeDetect} benchmark, a standard dataset for evaluating the generalization of AI-generated image detectors across diverse synthesis models \cite{ojha2023fakedetect}. The benchmark provides a unified evaluation protocol that enables consistent and fair comparison across detection approaches.
Following Wang \textit{et al.} \cite{Wang_2020_CVPR}, training is performed exclusively on ProGAN-generated images using multiple subsets with different generation configurations. This constrained training setup enables a rigorous assessment of cross-generator generalization to unseen synthesis pipelines.
The evaluation dataset spans \textbf{19 generative models}, covering a broad range of synthesis paradigms including GAN-based architectures, face-swapping techniques, low-level manipulation methods, perceptual-loss models, and diffusion-based generators. This diversity ensures that performance is assessed across heterogeneous generative distributions representative of real-world scenarios.

\subsection{Baselines}
Following established methodologies~\cite{tan2024c2pclipinjectingcategorycommon,10.1145/3746265.3759667}, we adopt accuracy (ACC) as the primary metric and report mean accuracy over all test subsets to assess cross-generator generalization.

We compare against representative methods including Co-occurrence~\cite{journals/corr/abs-1903-06836}, Freq-spec~\cite{9035107}, CNN-Spot~\cite{Wang_2020_CVPR}, FatchFor~\cite{10.1007/978-3-030-58574-7_7}, UniFD~\cite{ojha2023fakedetect}, LGrad~\cite{10203908}, F3Net~\cite{10.1007/978-3-030-58610-2_6}, FreqNet~\cite{10.1609/aaai.v38i5.28310}, NPR~\cite{Tan_2024_CVPR}, FatFormer~\cite{liu2024fat}, C2P-CLIP~\cite{tan2024c2pclipinjectingcategorycommon}, RINE~\cite{koutlis2024leveraging}, AntiFakePrompt~\cite{chang2024antifakepromptprompttunedvisionlanguagemodels}, and Bi-LORA~\cite{10.1111/exsy.13829}. Baseline results are taken from REVEAL~\cite{10.1145/3746265.3759667} under the same UniversalFakeDetect benchmark, ensuring fair comparison.

\subsection{Implementation Details.}

Our framework uses a CLIP ViT-L/14 backbone, with training restricted to the last two transformer layers to preserve pretrained representations while enabling task-specific adaptation. Spectral components (pixel-domain and feature-domain FFT blocks) remain fixed. Optimization uses Adam over 10 epochs (batch size 64) with an incremental learning strategy for progressive adaptation to evolving generative distributions.
Experiments run on an NVIDIA Quadro RTX~5000 GPU (16\,GB VRAM). Retrieval-augmented inference indexes embeddings in a local Milvus vector database, using cosine similarity to retrieve top-$k$ nearest neighbors with majority voting for final predictions.

\section{Results}
\label{RES}

\subsection{Comparison with State-of-the-Art}

We evaluate SPARK-IL against state-of-the-art detection methods on the UniversalFakeDetect benchmark, which encompasses 19 diverse generative models spanning GANs, deepfakes, low-level manipulations, perceptual-loss models, and diffusion-based generators.Table~\ref{tab:gen_comparison} shows that SPARK-IL outperforms all current methods with a mean accuracy (mAcc) of 94.60\%. This is a significant 13.22\% increase over a typical fine-tuned Vision Transformer baseline (81.5\%) and a notable improvement of 0.70\% over REVEAL (93.90\%).

\begingroup
\setlength{\textfloatsep}{6pt}
\setlength{\intextsep}{6pt}
\setlength{\floatsep}{6pt}
\setlength{\abovecaptionskip}{3pt}
\setlength{\belowcaptionskip}{3pt}

\begin{table}[h]
\centering
\caption{Comparison across GAN, DeepFakes, low-level, perceptual-loss, and diffusion-based generators.}
\scriptsize
\setlength{\tabcolsep}{2.6pt}
\renewcommand{\arraystretch}{1.15}

\resizebox{\textwidth}{!}{%
\begin{tabular}{l c
    c c c c c c
    c
    c c c c
    c
    c c c c
    c c
    c
    c}
\toprule

\multirow{2}{*}{Methods} & \multirow{2}{*}{Ref}
& \multicolumn{6}{c}{GAN}
& \multirow{2}{*}{DeepFakes}
& \multicolumn{4}{c}{Low level}
& \multirow{2}{*}{Perceptual loss}
& \multicolumn{4}{c}{LDM}
& \multicolumn{2}{c}{GLIDE}
& \multirow{2}{*}{DALL·E}
& \multirow{2}{*}{mAcc} \\

\cmidrule(lr){3-8}
\cmidrule(lr){10-13}
\cmidrule(lr){15-18}
\cmidrule(lr){19-20}

&
& \shortstack{Pro-\\GAN}
& \shortstack{Cycle-\\GAN}
& \shortstack{Big-\\GAN}
& \shortstack{Style-\\GAN}
& \shortstack{Gau-\\GAN}
& \shortstack{Star-\\GAN}
&
& SITD & SAN & CRN & IMLE
&
& \shortstack{200\\steps}
& \shortstack{200\\w/cfg}
& \shortstack{100\\steps}
& \shortstack{100\\27}
& \shortstack{50\\27}
& \shortstack{100\\10}
&
& \\

\midrule

Freq-spec        & WIFS2019    & 49.90 & 99.90 & 50.50 & 49.90 & 50.30 & 99.70 & 50.10 & 50.00 & 48.00 & 50.60 & 50.10 & 50.90 & 50.40 & 50.40 & 50.30 & 51.70 & 51.40 & 50.40 & 50.00 & 55.45 \\
Co-occurrence    & Electr. Img & 97.70 & 97.70 & 53.75 & 92.50 & 51.10 & 54.70 & 57.10 & 63.06 & 55.85 & 65.65 & 65.80 & 60.50 & 70.70 & 70.55 & 71.00 & 70.25 & 69.60 & 69.90 & 67.55 & 66.86 \\
CNN-Spot         & CVPR2020    & 99.99 & 85.20 & 70.20 & 85.70 & 78.95 & 91.70 & 53.47 & 66.67 & 48.69 & 86.31 & 86.26 & 60.07 & 54.03 & 54.96 & 54.14 & 60.78 & 63.80 & 65.66 & 55.58 & 69.58 \\
Patchfor         & ECCV2020    & 75.03 & 68.97 & 68.47 & 79.16 & 64.23 & 63.94 & 75.54 & 75.14 & 75.28 & 72.33 & 55.30 & 67.41 & 76.50 & 76.10 & 75.77 & 74.81 & 73.28 & 68.52 & 67.91 & 71.24 \\
F3Net            & ECCV2020    & 99.38 & 76.38 & 65.33 & 92.56 & 58.10 & 100.00& 63.48 & 54.17 & 47.26 & 51.47 & 51.47 & 69.20 & 68.15 & 75.35 & 68.80 & 81.65 & 83.25 & 83.05 & 66.30 & 71.33 \\
Bi-LoRA          & ICASSP2023  & 98.71 & 96.74 & 81.18 & 78.30 & 96.30 & 86.32 & 57.78 & 68.89 & 52.28 & 73.00 & 82.60 & 65.10 & 85.15 & 59.20 & 85.00 & 83.50 & 85.65 & 84.90 & 72.70 & 78.59 \\
LGrad            & CVPR2023    & 99.84 & 85.39 & 82.88 & 94.83 & 72.45 & 99.62 & 58.00 & 62.50 & 50.00 & 50.74 & 50.78 & 77.50 & 94.20 & 95.85 & 94.80 & 87.40 & 90.70 & 89.55 & 88.35 & 80.28 \\
UniFD            & CVPR2023    & 100.00& 98.50 & 94.50 & 82.00 & 99.50 & 97.00 & 66.60 & 63.00 & 57.50 & 59.50 & 72.00 & 70.03 & 94.19 & 73.76 & 94.36 & 79.07 & 79.85 & 78.14 & 86.78 & 81.38 \\
AntiFakePrompt   & CVPR2023    & 99.26 & 96.82 & 87.88 & 80.00 & 98.13 & 83.57 & 60.20 & 70.56 & 53.70 & 79.21 & 79.01 & 73.75 & 89.55 & 64.10 & 89.80 & 93.55 & 93.90 & 92.95 & 80.10 & 82.42 \\

REVEAL  & ACM 2025 
& \textbf{99.98} & 97.35 & 99.15 & 96.29 & 99.26 & \textbf{99.80}
& 93.49
& 95.00 & 64.16 & 95.34 & 95.34
& \textbf{68.90}
& 99.25 & \textbf{96.95} & 99.30 & 95.00
& 95.20 & 95.95
& 98.45
& 93.90 \\

\rowcolor{gray!15}
\textbf{SPARK-IL(ours)} & --
& 99.92 & \textbf{97.60} & \textbf{99.20} & \textbf{96.80} & \textbf{99.40} & 99.10
& \textbf{94.30}
& \textbf{95.80} & \textbf{65.20} & \textbf{96.10} & \textbf{96.00}
& 69.80
& \textbf{99.40} & 96.10 & \textbf{99.60} & \textbf{95.40}
& \textbf{95.90} & \textbf{96.20}
& \textbf{98.90}
& \textbf{94.60} \\

\bottomrule
\end{tabular}
}

\label{tab:gen_comparison}
\end{table}

% SPARK-IL achieves near-perfect performance on GAN-based generators, with accuracies of \textbf{99.92\%} on ProGAN and \textbf{99.20\%} on BigGAN. Strong performance is also observed on diffusion-based models, including \textbf{99.40\%} on LDM-200 and \textbf{98.90\%} on DALL·E. For challenging categories, SPARK-IL attains \textbf{95.80\%} accuracy on low-level manipulations (SITD) and \textbf{94.30\%} on DeepFakes. Performance on SAN remains lower (\textbf{65.20\%}), consistent with trends reported for all evaluated methods.

% In addition to quantitative results, Fig.~\ref{fig:tsne_grid_3x2} presents a qualitative analysis of the learned embedding space. The t-SNE visualizations show well-structured and separable embeddings for real and fake images across diverse generative models, including diffusion-based generators, indicating clear inter-class separation in the spectral embedding space.

SPARK-IL achieves \textbf{94.60\%} mAcc, outperforming REVEAL by \textbf{+0.7\%} and UniFD by \textbf{+13.2\%}. On GANs, accuracy exceeds 99\% (ProGAN: 99.92\%, BigGAN: 99.20\%). Diffusion models show consistent gains: LDM-200 reaches 99.40\% and DALL·E 98.90\%. Low-level manipulations (SITD: 95.80\%) and DeepFakes (94.30\%) improve by +0.8\% over prior work. SAN remains challenging at 65.20\%, consistent across all methods.
Figure ~\ref{fig:tsne_grid_3x2} visualizes the learned embedding space. The t-SNE projections reveal well-separated real/fake clusters across generator families, confirming effective inter-class discrimination in spectral space.

%The consistent performance gains across all generator families are particularly noteworthy. SPARK-IL achieves near-perfect accuracy (99.92\% on ProGAN and 99.20\% on BigGAN) on GAN-based models, demonstrating strong detection capabilities across a variety of adversarial architectures. Our approach maintains strong performance (99.40\% on LDM-200 and 98.90\% on DALL·E) for diffusion models, which are particularly difficult because of their high perceptual quality and statistical realism. Our dual spectral architecture contributes to this success by capturing frequency-domain artifacts that can persist even in highly realistic synthetic images. Spectral analysis identifies underlying statistical irregularities introduced during the generation process, in contrast to spatial-domain methods that could be tricked by visually appealing content.

%Additionally, SPARK-IL performs well in challenging categories like low-level manipulations (95.80\% on SITD) and DeepFakes (94.30\%). The relatively lower SAN score (65.20\%) aligns with trends observed in all evaluated techniques, indicating that this specific technique has intrinsic detection difficulties because of its subtle manipulation features. This restriction points to a potential area for future development, perhaps via feature engineering specific to SAN's distinct artifacts or specialized frequency band analysis.
\setlength{\textfloatsep}{6pt}  
\setlength{\intextsep}{6pt}      
\setlength{\floatsep}{6pt}       
\setlength{\abovecaptionskip}{3pt}
\setlength{\belowcaptionskip}{3pt}

\begin{figure*}[h]
    \centering

    % -------- Row 1 --------
    \includegraphics[width=0.32\textwidth]{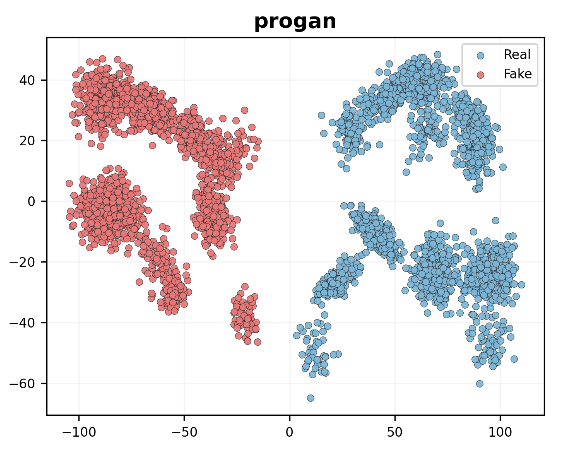}\hfill
    \includegraphics[width=0.32\textwidth]{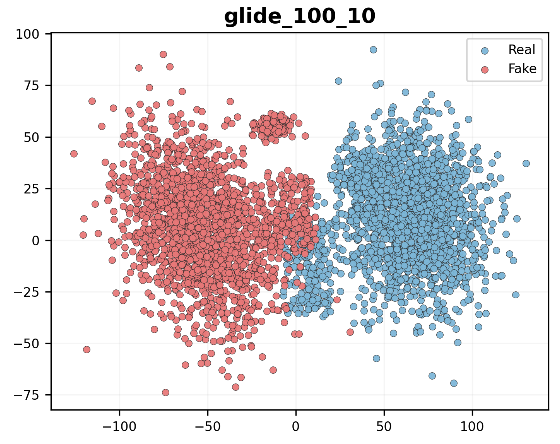}\hfill
    \includegraphics[width=0.32\textwidth]{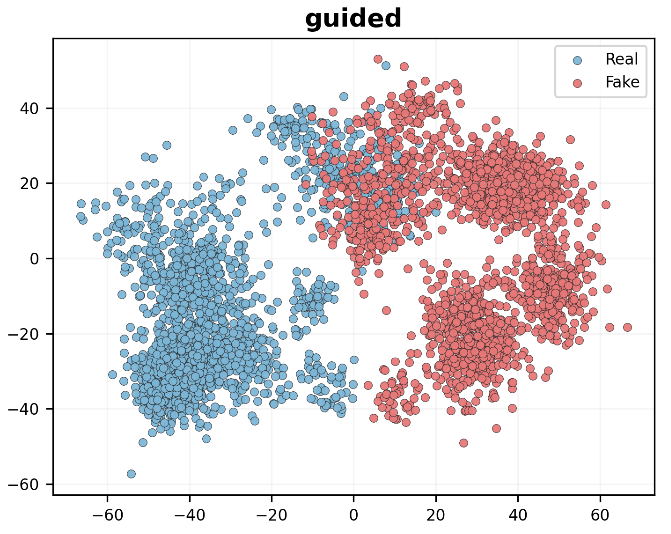}

    \vspace{0.5em}

    % -------- Row 2 --------
    \includegraphics[width=0.32\textwidth]{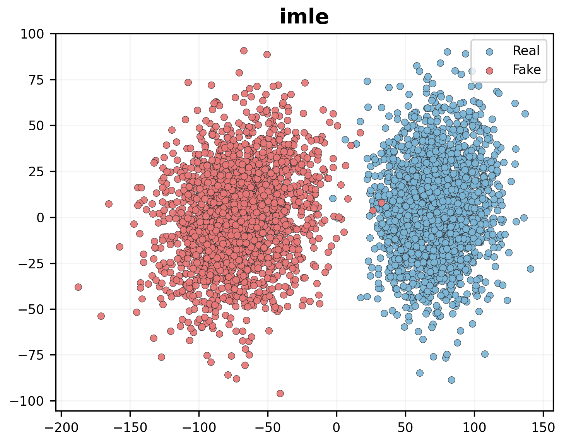}\hfill
    \includegraphics[width=0.32\textwidth]{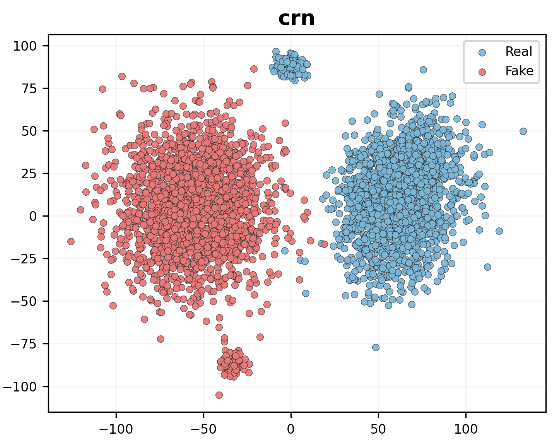}\hfill
    \includegraphics[width=0.32\textwidth]{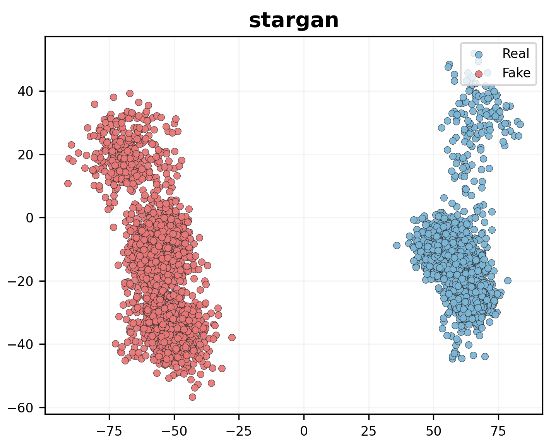}

    \caption{t-SNE visualization of SPARK-IL RAG embedding distributions for real (green) and fake (red) samples across different generative models.}
    \label{fig:tsne_grid_3x2}
\end{figure*}

\vspace{-5mm}
\subsection{Retrieval-Augmented Inference Analysis}

% In preamble (if not already):
% \usepackage{graphicx}
% \usepackage{subcaption}

A key component of SPARK-IL is its retrieval-augmented inference mechanism, which retrieves the top-$K$ most similar spectral embeddings from the database and determines predictions via majority voting. Table~\ref{tab:nshot_full} reports the impact of varying $K$ on detection performance.

With a single retrieved neighbor ($K{=}1$), SPARK-IL achieves 92.80\% mean accuracy. Performance improves consistently with increasing $K$, reaching 94.10\% at $K{=}5$ and 94.40\% at $K{=}10$. Beyond this point, gains saturate: $K{=}15$ yields 94.50\% and $K{=}20$ reaches 94.60\%, representing only marginal improvements over $K{=}10$. These results suggest that approximately 10--15 retrieved neighbors are sufficient to capture the discriminative structure of the spectral embedding space, with additional samples providing diminishing returns.

\begingroup
\setlength{\textfloatsep}{6pt}
\setlength{\intextsep}{6pt}
\setlength{\floatsep}{6pt}
\setlength{\abovecaptionskip}{3pt}
\setlength{\belowcaptionskip}{3pt}

\begin{table*}[h]
\centering
\caption{Effect of the number of retrieved neighbors (Nb shots) on SPARK-IL performance across all generator families on the UniversalFakeDetect benchmark. Performance consistently improves with more retrieved samples and stabilizes beyond 10–15 shots.}
\scriptsize
\setlength{\tabcolsep}{2.6pt}
\renewcommand{\arraystretch}{1.15}

\resizebox{\textwidth}{!}{%
\begin{tabular}{c
    c c c c c c
    c
    c c c c
    c
    c c c c
    c c
    c
    c}
\toprule

\textbf{Nb shots}
& \multicolumn{6}{c}{GAN}
& \textbf{DeepFakes}
& \multicolumn{4}{c}{Low level}
& \textbf{Perceptual}
& \multicolumn{4}{c}{LDM}
& \multicolumn{2}{c}{GLIDE}
& \textbf{DALL·E}
& \textbf{mAcc} \\

\cmidrule(lr){2-7}
\cmidrule(lr){9-12}
\cmidrule(lr){14-17}
\cmidrule(lr){18-19}

& ProGAN & CycleGAN & BigGAN & StyleGAN & GauGAN & StarGAN
& 
& SITD & SAN & CRN & IMLE
&
& 200 & 200 & 100 & 100
& 50 & 100
&
& \\

\midrule

1  
& 99.30 & 96.90 & 98.60 & 95.90 & 98.80 & 98.40
& 93.20
& 94.90 & 63.90 & 95.10 & 94.90
& 68.40
& 98.90 & 95.50 & 98.90 & 94.60
& 95.10 & 95.40
& 98.20
& 92.80 \\

3  
& 99.55 & 97.20 & 98.90 & 96.20 & 99.10 & 98.70
& 93.70
& 95.30 & 64.50 & 95.60 & 95.40
& 69.00
& 99.10 & 95.80 & 99.20 & 94.90
& 95.40 & 95.70
& 98.50
& 93.60 \\

5  
& 99.70 & 97.40 & 99.05 & 96.45 & 99.25 & 98.90
& 94.00
& 95.60 & 64.90 & 95.90 & 95.70
& 69.30
& 99.25 & 95.95 & 99.40 & 95.10
& 95.60 & 95.90
& 98.70
& 94.10 \\

10 
& 99.85 & 97.55 & 99.15 & 96.65 & 99.35 & 99.00
& 94.15
& 95.70 & 65.10 & 96.00 & 95.85
& 69.55
& 99.35 & 96.05 & 99.50 & 95.25
& 95.75 & 96.05
& 98.80
& 94.40 \\

15 
& 99.90 & 97.58 & 99.18 & 96.75 & 99.38 & 99.05
& 94.25
& 95.75 & 65.15 & 96.05 & 95.95
& 69.70
& 99.38 & 96.08 & 99.55 & 95.35
& 95.85 & 96.15
& 98.85
& 94.50 \\

\rowcolor{gray!15}
20 
& \textbf{99.92} & \textbf{97.60} & \textbf{99.20} & \textbf{96.80} & \textbf{99.40} & \textbf{99.10}
& \textbf{94.30}
& \textbf{95.80} & \textbf{65.20} & \textbf{96.10} & \textbf{96.00}
& \textbf{69.80}
& \textbf{99.40} & \textbf{96.10} & \textbf{99.60} & \textbf{95.40}
& \textbf{95.90} & \textbf{96.20}
& \textbf{98.90}
& \textbf{94.60} \\

\bottomrule
\end{tabular}
}

\label{tab:nshot_full}
\end{table*}

The results also exhibit generator-dependent behavior. GAN-based models perform strongly even at low shot counts (e.g., \textbf{98.40\%} on StarGAN at 1-shot), whereas more challenging techniques such as SAN benefit from additional retrieved samples, improving from \textbf{63.90\%} at 1-shot to \textbf{65.20\%} at 20 shots.

\subsection{Computational Efficiency Assessment}

SPARK-IL achieves state-of-the-art accuracy while maintaining competitive computational efficiency, table~\ref{tab:computational_comparison} summarizes computational costs. SPARK-IL comprises 315M parameters and 60.33 GFLOPs on $224{\times}224$ input, built on a partially frozen ViT-L/14 with only the last two blocks trained.

\begingroup
\setlength{\textfloatsep}{6pt}
\setlength{\intextsep}{6pt}
\setlength{\floatsep}{6pt}
\setlength{\abovecaptionskip}{3pt}
\setlength{\belowcaptionskip}{3pt}

\begin{table}
\centering
\scriptsize
\setlength{\tabcolsep}{3.5pt}
\renewcommand{\arraystretch}{1.15}
\setlength{\abovecaptionskip}{4pt}
\setlength{\belowcaptionskip}{4pt}
\captionsetup{singlelinecheck=false}
\caption{Comparison of computational complexity among recent methods, reporting the number of parameters and GFLOPs for a 224$\times$224 input.}
\label{tab:computational_comparison}
\begin{tabular}{l c c c}
\toprule
\textbf{Method} & \textbf{Backbone} & \textbf{Params (M)} & \textbf{GFLOPs} \\
\midrule
REVEAL & ViT-L/14 (partial) & 303.0 & 45.00 \\
C2P-CLIP & ViT-L/14 (partial) & 304.0 & 46.00 \\
RINE & ViT-L/14 & 434.0 & 52.00 \\
FatFormer & ViT-L/14 & 493.0 & 58.00 \\
\midrule
\rowcolor{gray!15}
\textbf{SPARK-IL (Ours)} & \textbf{ViT-L/14 (partial)} & \textbf{315.0} & \textbf{60.33} \\
\bottomrule
\end{tabular}
\end{table}

% \vspace{-8pt}
% \begin{table}[!h]
% \centering
% \scriptsize
% \setlength{\tabcolsep}{3.5pt}
% \renewcommand{\arraystretch}{1.2}
% \captionsetup{singlelinecheck=false} 
% \caption{Comparison of computational complexity among recent methods, reporting the number of parameters and GFLOPs for a 224×224 input.}
% \begin{tabular}{l c c c}
% \toprule
% \textbf{Method} & \textbf{Backbone} & \textbf{Params (M)} & \textbf{GFLOPs} \\
% \midrule
% REVEAL & ViT-L/14 (partial) & 303.0 & 45.00 \\
% C2P-CLIP & ViT-L/14 (partial) & 304.0 & 46.00 \\
% RINE & ViT-L/14 & 434.0 & 52.00 \\
% FatFormer & ViT-L/14 & 493.0 & 58.00 \\
% \midrule
% \rowcolor{gray!15}
% \textbf{SPARK-IL (Ours)} & \textbf{ViT-L/14 (partial)} & \textbf{315.0} & \textbf{60.33} \\
% \bottomrule
% \end{tabular}
% \label{tab:computational_comparison}
% \end{table}

Compared to REVEAL (303M), SPARK-IL uses +4\% parameters for +0.7\% accuracy—a favorable trade-off. Against larger models, SPARK-IL requires $1.56{\times}$ fewer parameters than FatFormer (493M) and $1.38{\times}$ fewer than RINE (434M), while achieving +3.7\% and +3.3\% higher accuracy respectively. The dual spectral architecture captures complementary information without excessive overhead, and retrieval-augmented inference expands effective capacity without adding trainable parameters.

\subsection{Ablation Study}

To quantify the contribution of each component in \textbf{SPARK-IL}, we conduct a systematic ablation study on the \textbf{UniversalFakeDetect} benchmark, with results summarized in Table~\ref{tab:ablation}. All variants follow the same \emph{incremental learning} protocol, where the Vision Transformer (ViT) backbone is \textbf{kept frozen} and only the \textbf{last two layers} are trained.

\begingroup
\setlength{\textfloatsep}{6pt}
\setlength{\intextsep}{6pt}
\setlength{\floatsep}{6pt}
\setlength{\abovecaptionskip}{3pt}
\setlength{\belowcaptionskip}{3pt}

\begin{table}[h]
\centering
\scriptsize
\setlength{\tabcolsep}{4.2pt}
\renewcommand{\arraystretch}{1.05}
\caption{Ablation study on the UniversalFakeDetect benchmark. 
The baseline corresponds to a standard Vision Transformer fine-tuned for synthetic image detection. 
Each row incrementally adds spectral modeling, classifier design, and retrieval-based reasoning.}
\begin{tabular}{l c c c c c}
\toprule
\textbf{Model Variant} 
& \textbf{Pixel FFT} 
& \textbf{Feature FFT} 
& \textbf{Retrieval + Voting} 
& \textbf{KAN} 
& \textbf{mAcc (\%)} \\
\midrule

(A) ViT Baseline                 & -- & -- & -- & -- & 81.5 \\
(B) + Pixel Spectral             & \checkmark & -- & -- & -- & 85.2 \\
(C) + Dual Spectral              & \checkmark & \checkmark & -- & -- & 88.7 \\
(D) + MLP Classifier             & \checkmark & \checkmark & -- & -- & 89.1 \\
(E) + KAN (w/o Retrieval)        & \checkmark & \checkmark & -- & \checkmark & 90.3 \\
\rowcolor{gray!15}
(F) \textbf{SPARK-IL (Full)}     & \checkmark & \checkmark & \checkmark & \checkmark & \textbf{94.6} \\

\bottomrule
\end{tabular}

\vspace{2pt}

\label{tab:ablation}
\end{table}

% \paragraph{Pixel-domain spectral modeling.}
Adding the pixel-level FFT branch (Variant B in Table~\ref{tab:ablation}) improves the mean accuracy to \textbf{85.2\%}, demonstrating the benefit of incorporating frequency information directly from raw RGB inputs.

% \paragraph{Dual spectral architecture.}
Introducing feature-domain spectral analysis by applying FFT to intermediate ViT embeddings (Variant C) further increases performance to \textbf{88.7\%}, indicating that pixel- and feature-level frequency representations provide complementary information.
% \paragraph{Projection head design.}
Replacing the linear classifier with a lightweight MLP projection head (Variant D) yields a consistent improvement to \textbf{89.1\%}, suggesting improved feature alignment in the spectral space.
% \paragraph{KAN-based frequency transformation.}
Employing Kolmogorov–Arnold Networks (KANs) for band-wise nonlinear transformations (Variant E) further boosts accuracy to \textbf{90.3\%}, highlighting the advantage of adaptive frequency-specific mappings under a frozen backbone.
% \paragraph{Retrieval-augmented inference.}
Enabling retrieval-based inference with majority voting over nearest neighbors results in the full \textbf{SPARK-IL} model (Variant F), achieving \textbf{94.6\%} mean accuracy. This corresponds to a \textbf{+4.3\%} improvement over Variant E and represents the largest gain observed in the ablation study.
% \paragraph{Overall analysis.}
The results in Table~\ref{tab:ablation} confirm that each component contributes meaningfully to performance, with \emph{dual spectral modeling} and \emph{retrieval-augmented inference} providing the most substantial improvements.

\section{Discussion}
\label{sec:discussion}
\vspace{-2mm}

The experimental results provide empirical support for the design choices underlying SPARK-IL and reveal both the strengths and limitations of spectral analysis for cross-generator deepfake detection.
\vspace{-5mm}
\subsection{Spectral Analysis Architecture}

The dual-path spectral architecture applies FFT at two complementary abstraction levels, each targeting distinct generation artifacts. Pixel-domain FFT reveals low-level artifacts such as periodic patterns from upsampling and quantization noise that are often latent in the spatial domain but explicit in the frequency spectrum. Feature-domain FFT operates on ViT embeddings to capture semantic-level inconsistencies where learned spectral structures deviate from those of natural images. Cross-attention selectively fuses these spectral views by aligning pixel-level frequency cues with semantically relevant feature-level components. As quantified in Table~\ref{tab:ablation}, pixel-domain FFT yields a +3.7\% gain over the baseline, while adding feature-domain FFT provides an additional +3.5\%, confirming that the two representations encode complementary artifact signatures.

Category-wise results in Table~\ref{tab:gen_comparison} show that generator families exhibit characteristic spectral profiles. GANs achieve detection rates above 96\%, with ProGAN, BigGAN, and GauGAN exceeding 99\%, reflecting fingerprints from transposed convolutions and progressive upsampling. Diffusion models distribute artifacts more broadly due to iterative denoising. Strong performance on LDM, GLIDE, and DALL·E demonstrates that multi-band decomposition generalizes across distinct generation paradigms.

The four-band partition supports band-specific processing: low-frequency bands encode global structure and illumination, while high-frequency bands capture texture irregularities and aliasing. KAN addresses the differing statistics of each band by learning adaptive transformations via spline-based activations, contributing +1.2\% over MLP layers (Table~\ref{tab:ablation}).

\subsection{Retrieval and Incremental Learning}

Retrieval-augmented classification addresses a key limitation of parametric classifiers: fixed decision boundaries generalize poorly when test embeddings lie outside the training distribution, a common scenario in deepfake detection as new generators emerge. The retrieval mechanism replaces a global decision rule with a local one by aggregating labels from the $k$ nearest stored examples, enabling adaptation to the query distribution without parameter updates. Its effectiveness relies on the quality of the embedding space; the dual-path spectral representation produces compact embeddings where similar generators cluster closely, supporting accurate nearest-neighbor classification. As shown in Table~\ref{tab:ablation}, retrieval yields the largest single performance gain (+4.3\%), indicating that non-parametric evidence complements learned parametric features.

Incremental learning is enabled by the combination of weight consolidation and database expansion. Elastic weight consolidation preserves band-specific transformations learned from earlier generators, reducing catastrophic forgetting, while embeddings from new generators are added to the retrieval database without overwriting existing entries. This design supports cumulative knowledge acquisition, with learned transformations retained in the model and exemplar knowledge retained in the database. Table~\ref{tab:nshot_full} confirms this behavior: accuracy increases monotonically from 1-shot to 20-shot configurations (+1.8\%) without degrading performance on previously seen techniques, demonstrating effective adaptation to new generators.

\vspace{-2mm}
\section{Conclusion and Future Work}
\label{sec:conc}
\vspace{-2mm}

We presented SPARK-IL, a retrieval-augmented framework for incremental deepfake detection that combines dual-path spectral analysis with KAN-based feature transformation. By applying FFT at both pixel and feature levels, SPARK-IL captures complementary artifact signatures ranging from low-level generation noise to higher-level semantic inconsistencies, which are effectively fused through cross-attention. Experiments on the UniversalFakeDetect benchmark demonstrate state-of-the-art performance, highlighting the stability of frequency-domain representations for cross-generator generalization. Retrieval-augmented inference provides non-parametric correction when classifier confidence is limited and yields the largest individual performance gain, while new generators are integrated through simple database expansion without retraining or degradation of previously seen techniques. Despite these strengths, several limitations remain. Global FFT is less effective for spatially localized artifacts, motivating future work on spatially aware or patch-based retrieval. In addition, adaptive frequency partitioning and multi-scale spectral analysis may further improve robustness as generative models continue to evolve.\\

{\noindent \bf Acknowledgments}:  
This work has been partially supported by the project PCI2022-134990-2 (MARTINI) of the CHISTERA IV Cofund 2021 program. Abdenour Hadid is funded by TotalEnergies collaboration agreement with Sorbonne University Abu Dhabi.

%\bibliographystyle{unsrtnat}
 %\bibliographystyle{splncs04}
 %\bibliography{references.bib}

 \bibliographystyle{splncs04}
 \setlength{\bibsep}{1pt}
 \bibliography{references}

\begin{thebibliography}{10}
\providecommand{\url}[1]{\texttt{#1}}
\providecommand{\urlprefix}{URL }
\providecommand{\doi}[1]{https://doi.org/#1}

\bibitem{bonomo2025visualragexpandingmllm}
Bonomo, M., Bianco, S.: Visual rag: Expanding mllm visual knowledge without fine-tuning. arXiv preprint arXiv:2501.10834  (2025)

\bibitem{Caffagni2024WikiLLaVA}
Caffagni, D., Cocchi, F., Moratelli, N., Sarto, S., Cornia, M., Baraldi, L., Cucchiara, R.: Wiki-llava: Hierarchical retrieval-augmented generation for multimodal llms. In: Proceedings of the IEEE/CVF Conference on Computer Vision and Pattern Recognition Workshops. pp. 1818--1826 (2024)

\bibitem{10.1007/978-3-030-58574-7_7}
Chai, L., Bau, D., Lim, S.N., Isola, P.: What makes fake images detectable? understanding properties that generalize. In: European Conference on Computer Vision (2020)

\bibitem{chang2024antifakepromptprompttunedvisionlanguagemodels}
Chang, Y.M., Yeh, C., Chiu, W.C., Yu, N.: Antifakeprompt: Prompt-tuned vision-language models are fake image detectors. arXiv preprint arXiv:2310.17419  (2024)

\bibitem{10860248}
Chen, Y., Yashtini, M.: Detecting ai generated images through texture and frequency analysis of patches. In: International Conference on Artificial Intelligence, Virtual Reality and Visualization (2024)

\bibitem{cozzolino2023raising}
Cozzolino, D., Poggi, G., Corvi, R., Nie{\ss}ner, M., Verdoliva, L.: Raising the bar of ai-generated image detection with clip. In: Proceedings of the IEEE/CVF Conference on Computer Vision and Pattern Recognition Workshops (2024)

\bibitem{dhariwal2021diffusion}
Dhariwal, P., Nichol, A.Q.: Diffusion models beat gans on image synthesis. In: Advances in Neural Information Processing Systems (2021)

\bibitem{NIPS2014_f033ed80}
Goodfellow, I.J., Pouget-Abadie, J., Mirza, M., Xu, B., Warde-Farley, D., Ozair, S., Courville, A., Bengio, Y.: Generative adversarial nets. In: Advances in Neural Information Processing Systems (2014)

\bibitem{NEURIPS2024_efaf1c97}
He, X., Tian, Y., Sun, Y., Chawla, N.V., Laurent, T., LeCun, Y., Bresson, X., Hooi, B.: G-retriever: Retrieval-augmented generation for textual graph understanding and question answering. In: Advances in Neural Information Processing Systems (2024)

\bibitem{karras2018progressive}
Karras, T., Aila, T., Laine, S., Lehtinen, J.: Progressive growing of gans for improved quality, stability, and variation. In: International Conference on Learning Representations (2018)

\bibitem{10.1111/exsy.13829}
Ke{\"i}ta, M., Hamidouche, W., Eutamene, H.B., Taleb-Ahmed, A., Camacho, D., Hadid, A.: Bi-lora: A vision-language approach for synthetic image detection. Expert Systems  (2025)

\bibitem{10.1145/3746265.3759667}
Keita, M., Hamidouche, W., Eutamene, H.B., Taleb-Ahmed, A., Hadid, A.: Reveal: A retrieval-augmented generation approach for contextual identification of synthetic visual content. In: Proceedings of the Deepfake Forensics Workshop (2025)

\bibitem{RAVID}
Keita, M., Hamidouche, W., Eutamene, H.B., Taleb-Ahmed, A., Hadid, A.: Ravid: Retrieval-augmented visual detection. arXiv preprint arXiv:2508.03967  (2025)

\bibitem{doi:10.1073/pnas.1611835114}
Kirkpatrick, J., Pascanu, R., Rabinowitz, N., Veness, J., Desjardins, G., Rusu, A.A., Milan, K., Quan, J., Ramalho, T., Grabska-Barwinska, A., Hassabis, D., Clopath, C., Kumaran, D., Hadsell, R.: Overcoming catastrophic forgetting in neural networks. Proceedings of the National Academy of Sciences  (2017)

\bibitem{koutlis2024leveraging}
Koutlis, C., Papadopoulos, S.: Leveraging representations from intermediate encoder-blocks for synthetic image detection. In: European Conference on Computer Vision (2024)

\bibitem{lewis2020retrieval}
Lewis, P., Oguz, B., Rinott, R., Riedel, S., Rockt{\"a}schel, T., et~al.: Retrieval-augmented generation for knowledge-intensive natural language processing. Advances in Neural Information Processing Systems  (2020)

\bibitem{liu2024fat}
Liu, H., Tan, Z., Tan, C., Wei, Y., Wang, J., Zhao, Y.: Forgery-aware adaptive transformer for generalizable synthetic image detection. In: Proceedings of the IEEE/CVF Conference on Computer Vision and Pattern Recognition (2024)

\bibitem{long2025retrievalaugmentedvisualquestionanswering}
Long, X., Ma, Z., Hua, E., Zhang, K., Qi, B., Zhou, B.: Retrieval-augmented visual question answering via built-in autoregressive search engines. arXiv preprint arXiv:2502.16641  (2025)

\bibitem{journals/corr/abs-1903-06836}
Nataraj, L., Mohammed, T.M., Manjunath, B.S., Chandrasekaran, S., Flenner, A., Bappy, J.H., Roy-Chowdhury, A.K.: Detecting gan generated fake images using co-occurrence matrices. arXiv preprint arXiv:1903.06836  (2019)

\bibitem{nichol2022glide}
Nichol, A.Q., Dhariwal, P., Ramesh, A., Shyam, P., Mishkin, P., McGrew, B., Sutskever, I., Chen, M.: Glide: Towards photorealistic image generation and editing with text-guided diffusion models. In: International Conference on Machine Learning (2022)

\bibitem{ojha2023fakedetect}
Ojha, U., Li, Y., Lee, Y.J.: Towards universal fake image detectors that generalize across generative models. In: Proceedings of the IEEE/CVF Conference on Computer Vision and Pattern Recognition (2023)

\bibitem{openai2023dalle3}
{OpenAI}: Dall·e 3: Improving image generation with better captions. Tech. rep. (2023)

\bibitem{10.1007/978-3-030-58610-2_6}
Qian, Y., Yin, G., Sheng, L., Chen, Z., Shao, J.: Thinking in frequency: Face forgery detection by mining frequency-aware clues. In: European Conference on Computer Vision (2020)

\bibitem{ramesh2021zero}
Ramesh, A., Pavlov, M., Goh, G., Gray, S., Voss, C., Radford, A., Chen, M., Sutskever, I.: Zero-shot text-to-image generation. In: International Conference on Machine Learning (2021)

\bibitem{Rombach_2022_CVPR}
Rombach, R., Blattmann, A., Lorenz, D., Esser, P., Ommer, B.: High-resolution image synthesis with latent diffusion models. In: Proceedings of the IEEE/CVF Conference on Computer Vision and Pattern Recognition (2022)

\bibitem{saharia2022photorealistic}
Saharia, C., Chan, W., Saxena, S., Li, L., Whang, J., Denton, E., Ghasemipour, S.K., Gontijo-Lopes, R., Ayan, B.K., Salimans, T., Ho, J., Fleet, D.J., Norouzi, M.: Photorealistic text-to-image diffusion models with deep language understanding. In: Advances in Neural Information Processing Systems (2022)

\bibitem{tan2024c2pclipinjectingcategorycommon}
Tan, C., Tao, R., Liu, H., Gu, G., Wu, B., Zhao, Y., Wei, Y.: C2p-clip: Injecting category common prompt in clip to enhance generalization in deepfake detection. arXiv preprint arXiv:2408.09647  (2024)

\bibitem{10.1609/aaai.v38i5.28310}
Tan, C., Zhao, Y., Wei, S., Gu, G., Liu, P., Wei, Y.: Frequency-aware deepfake detection: Improving generalizability through frequency space domain learning. In: Proceedings of the AAAI Conference on Artificial Intelligence (2024)

\bibitem{Tan_2024_CVPR}
Tan, C., Zhao, Y., Wei, S., Gu, G., Liu, P., Wei, Y.: Rethinking the up-sampling operations in cnn-based generative network for generalizable deepfake detection. In: Proceedings of the IEEE/CVF Conference on Computer Vision and Pattern Recognition (2024)

\bibitem{10203908}
Tan, C., Zhao, Y., Wei, S., Gu, G., Wei, Y.: Learning on gradients: Generalized artifacts representation for gan-generated images detection. In: Proceedings of the IEEE/CVF Conference on Computer Vision and Pattern Recognition (2023)

\bibitem{tao2023galip}
Tao, M., Bao, B.K., Tang, H., Xu, C.: Galip: Generative adversarial clips for text-to-image synthesis. In: Proceedings of the IEEE/CVF Conference on Computer Vision and Pattern Recognition (2023)

\bibitem{Wang_2020_CVPR}
Wang, S.Y., Wang, O., Zhang, R., Owens, A., Efros, A.A.: Cnn-generated images are surprisingly easy to spot... for now. In: Proceedings of the IEEE/CVF Conference on Computer Vision and Pattern Recognition (2020)

\bibitem{wang2023dire}
Wang, Z., Guo, J., Li, R., Hu, R., Zhou, H., Huang, R., Chen, Y.: Dire: Diffusion reconstruction error for diffusion-generated image detection. In: Proceedings of the IEEE/CVF International Conference on Computer Vision (2023)

\bibitem{Xu2024UFOGen}
Xu, Y., Zhao, Y., Xiao, Z., Hou, T.: Ufogen: You forward once large-scale text-to-image generation via diffusion gans. In: Proceedings of the IEEE/CVF Conference on Computer Vision and Pattern Recognition (2024)

\bibitem{9035107}
Zhang, X., Karaman, S., Chang, S.F.: Detecting and simulating artifacts in gan fake images. In: IEEE International Workshop on Information Forensics and Security (2019)

\end{thebibliography}

\end{document}